\documentclass[letterpaper, 10 pt, conference]{ieeeconf}

\usepackage{PackageHeading}
\usepackage{subfiles}
\usepackage{caption}
\usepackage{subcaption}
\usepackage[absolute,overlay]{textpos}
\title{\Large \bf
Visual Perception System for Autonomous Driving
}
\author{Qi Zhang$^\dagger$, Siyuan Gou$^\dagger$, and Wenbin Li$^\dagger$% <-this % stops a space
\thanks{$^\dagger$Department of Computer Science,
        University of Bath, UK, \{\protect\url{qz727, sg2655, w.li}\}\protect\url{@bath.ac.uk}}%
}

\begin{document}

\maketitle
\thispagestyle{empty}
\pagestyle{empty}

\begin{abstract}

The recent surge in interest in autonomous driving stems from its rapidly developing capacity to enhance safety, efficiency, and convenience. A pivotal aspect of autonomous driving technology is its perceptual systems, where core algorithms have yielded more precise algorithms applicable to autonomous driving, including vision-based Simultaneous Localization and Mapping (SLAMs), object detection, and tracking algorithms. This work introduces a visual-based perception system for autonomous driving that integrates trajectory tracking and prediction of moving objects to prevent collisions, while addressing autonomous driving's localization and mapping requirements. The system leverages motion cues from pedestrians to monitor and forecast their movements and simultaneously maps the environment. This integrated approach resolves camera localization and the tracking of other moving objects in the scene, subsequently generating a sparse map to facilitate vehicle navigation. The performance, efficiency, and resilience of this approach are substantiated through comprehensive evaluations of both simulated and real-world datasets.

\end{abstract}

\section{Introduction}

Perception, including object recognition and prediction, along with Visual SLAM (V-SLAM) systems, plays a crucial role in autonomous driving systems.
In particular, the reconstruction of a map of the surrounding environment, identification of moving objects, and prediction of their trajectories, are essential in promoting the safety of operating an autonomous vehicle in the real world.
Although different perception solutions may rely on various sensors, such as LiDAR, GPS, RGB-D cameras, etc., the visual only method remains the essential approach due to its low cost nature.

Autonomous driving under unstructured scenarios is challenging for V-SLAM systems, as they are required to handle dynamic objects on the road, such as pedestrians, and other moving vehicles, etc., using only visual images.
In a typical solution, visual SLAM systems may focus on improving odometry accuracy by removing moving objects as outliers in dynamic environments, as demonstrated in DynaSLAM \cite{bescos2018dynaslam}.
Nevertheless, some studies have attempted to detect and take moving objects into account together with the static environment structure using VDO-SLAM \cite{zhang2020vdo} and Cubeslam \cite{yang2019cubeslam}.
While these methods are limited by a lack of using the semantic clues from moving objects, DynaSLAM2 \cite{bescos2021dynaslam} has integrated object detection and tracking with the SLAM system to address this issue.

The driving scenario may be complex in real life due to the frequent presence of a number of moving objects. There is still a gap in the literature on the prediction of moving object trajectories along with a dynamic SLAM solution which is supposed to improve collision avoidance when autonomous vehicles are driving around. In this work, we propose a V-SLAM system that integrates moving object tracking and trajectory prediction to provide a perception solution for autonomous driving, which benefits not only localization and mapping quality but also the sufficiency of motion planning and collision avoidance.

Visual cues have gained popularity within the community for extracting rich semantic information from moving objects for dynamic V-SLAM systems.
Several data-driven approaches, including QD-3DT \cite{hu2022monocular}, TripletTrack \cite{marinello2022triplettrack}, and CC-3DT \cite{fischer2022cc}, are capable of detecting various moving objects, such as pedestrians, vehicles, and cyclists, even in complex road scenarios.
This prompted us to enhance our perception system by incorporating both 2D and 3D clues from moving objects.
However, the visual information of moving objects often introduces jitter into the system, especially when a large number of fast-moving objects are present in the view.
Those issues may lead to failures when detecting objects in general.
This occurs because the 3D information of moving objects is discretely learned from frame to frame. This nonlinear nature may lose information and give noisy and discontinuous training outcomes.
Our proposed method effectively addresses such issues through a lightweight trajectory prediction approach, enabling reliable and precise predictions.

\begin{figure}[tb]
     \centering
     \begin{subfigure}[thpb]{0.46\textwidth}
         \centering
         \includegraphics[width=\textwidth]{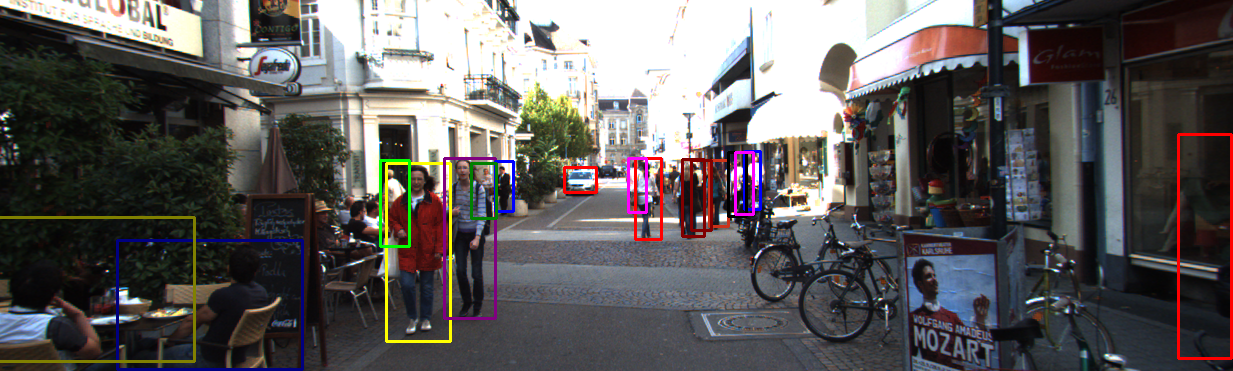}
         \label{final2d}
     \end{subfigure}
     \begin{subfigure}[thpb]{0.42\textwidth}
         \centering
         \includegraphics[width=\textwidth]{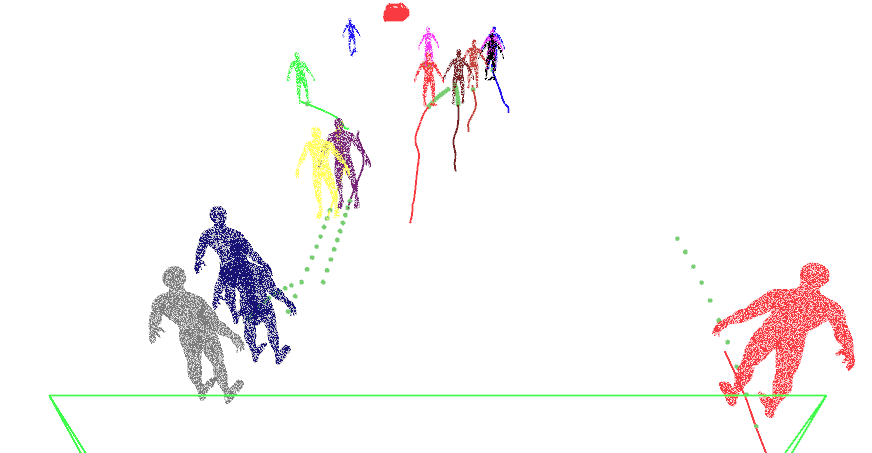}
         \label{final3d}
     \end{subfigure}
      \caption{A brief result of our perception system. The top image is an RGB frame. The bottom image shows the perception result constructed for this frame, including the reconstruction of moving objects classified with colors, as well as their predicted trajectories indicated by green dashes.}
      \label{result}
\end{figure}

To achieve effective collision avoidance, the system requires comprehensive access to scene information and the ability to predict the motion trajectories of moving objects. However, many current studies make use of the models trained by top-view street images with fixed camera views, focusing solely on predicting pedestrian trajectories in similar road scenarios \cite{zhang2019sr} \cite{peng2022srai} \cite{zhou2021ast} \cite{bae2022non} \cite{li2019grip}  \cite{li2019grip++}.
In contrast, autonomous vehicle systems may employ moving cameras and navigate through a wide range of road conditions. This may challenge existing approaches due to potential difficulties in generalization. Furthermore, some autonomous vehicles rely solely on front-facing camera information, which may not align with many of these existing approach requirements.
Thus, we developed an approach to bridge the gap in research between current visual perception systems for autonomous driving, and V-SLAM systems with real-time pedestrian and vehicle trajectory prediction.

In this paper, we introduce a stereo SLAM system that is capable of tracking and recognizing dynamic objects as shown in Fig. \ref{result}. Furthermore, we present a method that eliminates the noise of the 3D information of machine learning-based objects when integrating them into the SLAM system. In this background, we propose a lightweight approach for predicting the future trajectories of moving objects. This approach enables us to improve the tracking system by incorporating predicted trajectories.

\begin{itemize}
    \item A visual-based perception system for autonomous driving scenarios incorporates a dynamic SLAM algorithm (that removes the moving objects as outliers), together with moving object trajectory optimization and prediction. This can take into account any 2D and 3D bounding box detection method.
    \item A concise architecture for predicting trajectories of multiple moving objects, including both pedestrians and vehicles, simultaneously for each frame. This approach can also re-track object trajectories in situations where the data driven tracking method fails in a short period, leading to more accurate and robust tracking.
    \item A method solves the jitter problem when applying neural network-based object tracking methods to real-time SLAM systems, enabling more accurate trajectory predictions.
\end{itemize}

\section{Related Work}

Typically, Visual SLAM systems assume that the environment is static, such as the ORB-SLAM series \cite{mur2015orb} \cite{mur2017orb}.
However, dynamic environments pose a significant challenge to the process of feature association across views or frames.
In current research focused on SLAM systems in dynamic scenes, features in the dynamic region are often treated as outliers and subsequently removed to improve pose estimation accuracy \cite{bescos2018dynaslam} \cite{xiao2019dynamic}.
Some alternative approaches have shifted their focus towards utilizing neural networks for the semantic recognition of moving objects \cite{bescos2021dynaslam} \cite{yu2018ds} and leveraging 3D object information to optimize camera poses \cite{yang2019cubeslam}.
Additionally, dense optical flow algorithms, as shown in studies like \cite{zhang2020vdo} and \cite{wadud2022dyob}, are employed to distinguish between moving from static ones and to make an effort to track these objects.
Nevertheless, for autonomous driving system, not only localization and mapping in dynamic scenario is essential, but also predicting object trajectories hold significant importance, and there is a notable gap in research within this particular area.

The research on object detection and tracking has been extensive in 2D images and videos with well-established methods, such as YOLO series \cite{ge2021yolox}, Perma-track \cite{tokmakov2021learning}, and Observation-Centric SORT \cite{cao2022observation}, providing robust solutions for tracking complex objects in image sequences.
Tracking 3D objects from 2D images is a challenging task that requires recognizing objects from 2D images and extracting directional, volumetric, rotational, and translational information from a sequence of images.
To address this challenge, \cite{hu2022monocular} takes into account Long-Short Term Memory (LSTM) motion extrapolation to integrate previous 3D information with current information when using only visual images.
However, when applying the detected position to an online SLAM system, there are always jittering issues with sequential results. Therefore, we propose a lightweight method to eliminate jittering when using our SLAM system.

To predict the motion trajectories of pedestrians, a popular solution is interleaving an LSTM network that follows a message-passing mechanism to effectively extract social influence from neighbors, including SR-LSTM \cite{zhang2019sr} and SRAI-LSTM \cite{peng2022srai}, as well as methods that employ other network structures, such as \cite{zhou2021ast} and \cite{bae2022non}.
Such methodology may also support the prediction of vehicle trajectories as demonstrated in Grip \cite{li2019grip} and Grip++ \cite{li2019grip++}.
However, these methods are often trained static top-view cameras, and may need intensive computational resources, making them cost-prohibitive for autonomous driving systems.
To address this issue, CAPOs \cite{mcallister2022control} designed an autonomous driving system within the simulated environment implemented using CARLA \cite{dosovitskiy2017carla}.
Although it combines a deep-learning method to predict the movement of objects, allowing the autonomous car to efficiently reduce collision risks within a 3-second prediction window, it does not offer localization and mapping capabilities.
As a result, the system exhibits a notable drawback in terms of a high Average Displacement Error (ADE).

\section{Method}

\subsection{Perception system with SLAM system}

\begin{figure*}
     \centering

    \vspace{5pt}

     \includegraphics[width=0.9\textwidth]{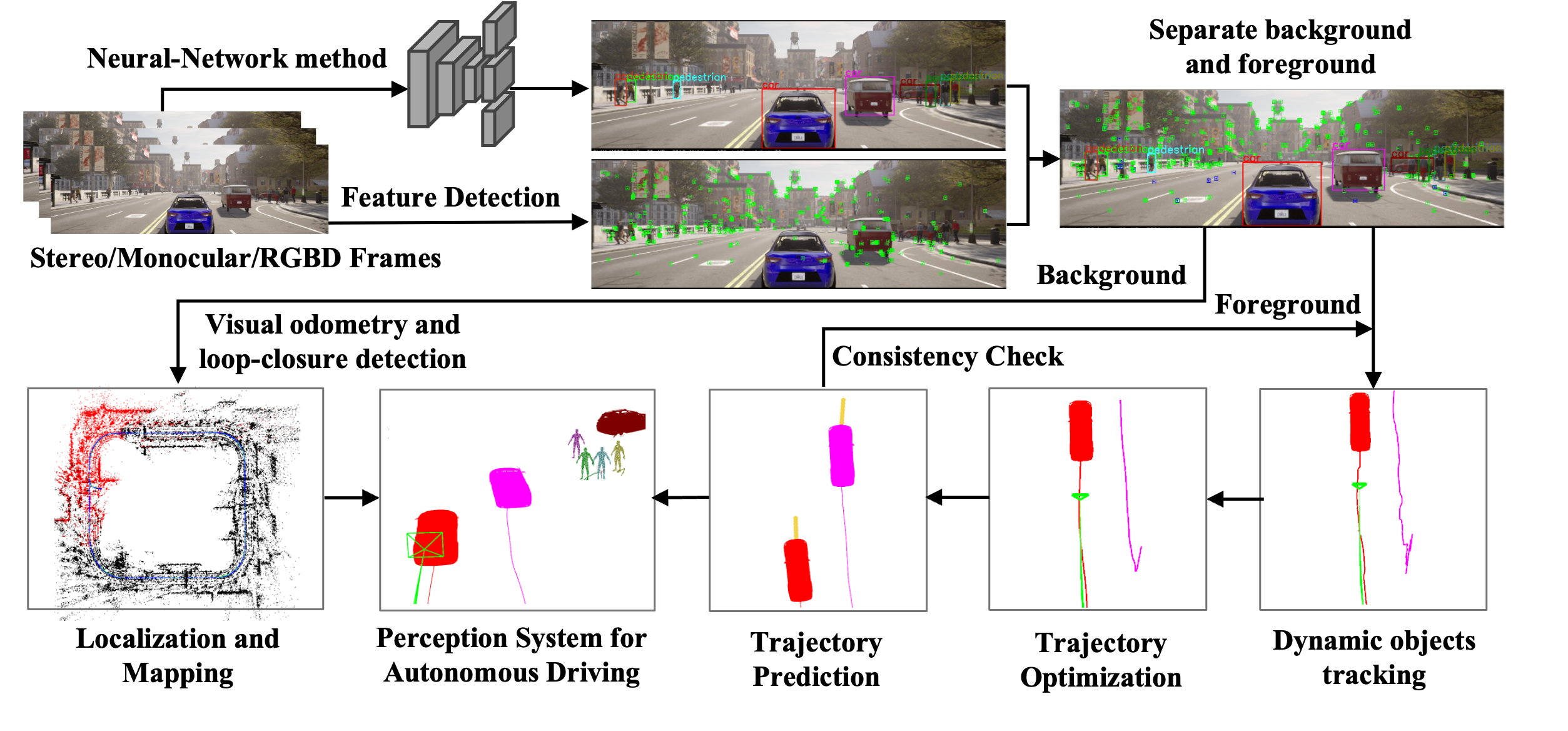}
     \caption{The overview of our system. We first employ data driven 3D bounding box detection method onto the frame image. We then extract background features and match them to create a sparse 3D map based on ORB-SLAM2 \cite{mur2017orb}. Simultaneously, we estimate the trajectories of the moving objects and predict their future paths over a short time period.}
    \label{overview}
\end{figure*}

As shown in Fig. \ref{overview}, we propose a perception that integrates an online SLAM system tightly, object tracking, and trajectory prediction.
While processing moving objects (such as vehicles, pedestrians, etc.), we track their positions when a new frame $F$ arrives.
We achieve a set of objects $\left\{O_N, \ldots, O_M\right\} = \left\{O_i\right\}^M_{i = N}$ for each frame where the subscripts are associated with the object id $\left\{N, \ldots, M\right\}$.
The centroid of an object is denoted as $\left\{P_N, \ldots, P_M\right\} = \left\{P_i\right\}_{i=N}^{M}$, where $P_i=\left(x_i, y_i, z_i\right) $ and heading directions of the objects (the Euler angle with axis $z$) are presented as $\left\{D_N, \ldots, D_M\right\} = \left\{D_i\right\}_{i=N}^{M}$, where $D_i \in (-\pi, \pi)$.

In addition to moving object tracking, we simultaneously predict the trajectories of these objects based on their real-time positions.
Composing the central points of each frame, we can obtain a set of moving objects trajectories as $\left\{S_N, \ldots, S_M\right\}= \left\{S_i\right\}_{i=N}^{M}$.
Assuming the current frame is $F_k$, we have the object trajectory $S_i = \{ P_{i}^{j}\}_{j=a}^{k}$, where $a$ is the object appeared at the $a$-th frame.
Moreover, the object trajectory prediction in the future time interval $T_{p}$ is defined as $C_{p,i} = \{ P_{i}^{j}\}_{j=k}^{k+b}$, where $b = T_{p} / {\Delta T}$ is the number of future trajectory prediction points.
% current k in (past_start_a, future_end_b)
% utilized_history_points h < k-a
We then further utilize $\left\{C_{p,i}\right\}_{i=N}^M$ for checking the consistency of the object tracking results.

\begin{algorithm}[tp]
\caption{Online Trajectory Prediction}\label{alg:overview}
\renewcommand{\algorithmicrequire}{\textbf{Input:}}
 \renewcommand{\algorithmicensure}{\textbf{Output:}}
\begin{algorithmic}[1]
\Require $\{S_i\}_{i=N}^M$
\While{Detect new frame $F_k$}
\For{$S_i $ in $\{S_i\}_{i=N}^M$}
\State Consistency check (Sec. \ref{consistency})
\State Kalman filter on $S_i $ (Sec. \ref{Initialization})
\If{$ |S_i| < \tau_{KF}$    }
    \State Initialization trajectory (Sec. \ref{Initialization})
    \State Update $S_i $
\ElsIf{$ |S_i| > \tau_{KF}$}
    \State Local trajectory optimization (Sec. \ref{TrajectoriesBA})
    \State Update $S_i $
    \State Calculate $P_{s, i}$ $P_{c, i}$ $P_{e, i} $ $C_{p,i}$ (Sec. \ref{Prediction})
\EndIf
\EndFor
\EndWhile
\Ensure $\{C_{p,i}\}_{i=N}^M$
\end{algorithmic}
\end{algorithm}

\subsection{Heading-based Moving Objects Trajectories Prediction}
\label{Prediction}

Although we can track the moving objects separately using neural-network based method, it is difficult to predict the trajectories of the moving objects in real time, especially when a large number of objects are observed in each frame. Thus, we propose a light-weight, fast-moving object trajectory prediction algorithm based on the head direction vector. The details are shown in Algorithm \ref{alg:overview}.

To simplify the problem and align with the nature of autonomous driving, we choose not to factor in the $z$ axis (vertical direction) when predicting the trajectory.
This decision is influenced by the small displacement observed along the $z$ axis and other relevant considerations for autonomous driving.
We assume that real-world motion patterns of moving objects, such as pedestrians and vehicles, have motion curves that are independent of each other. Although these curves may intersect, our focus is on ensuring that our vehicle does not collide with pedestrians or other vehicles, without concern for collisions between those external entities themselves.
In this case, we identify patterns and make predictions about future trajectories by analyzing a sequence of past object positions.
The object trajectory prediction model is depicted in Fig. \ref{fig_bazier}, where the curve $C_p$ represents a section of the trajectory of interest.
Considering short time intervals, it is often observed that the target trajectories do not exhibit significant curvature in various directions.
To approximate such trajectories, we employ a quadratic Bezier Curve, which necessitates three points: $P_s$, $P_e$, and $P_c$ (with $P_s$ being a known point).
To calculate the position of $P_c$, we use the head direction vector $\mathbf{V_h}$ and the endpoint $P_e$.
We then determine the position of $P_e$, we rely on the predicted vector $\mathbf{V_p}$ and the trajectory distance $d_{se} = \overline{P_s P_e}$.

\subsubsection{\textbf{Head Direction Vector}}
The head direction $D_i$ obtained from the per-frame object detection is converted into unit direction (normal) vectors $\mathbf{\hat{d}}_i$.
However, the directions from the previous learning-based detection are always noisy an unreliable. We thus calculate the instantaneous tangent vector $\mathbf{a}_{i} = \overrightarrow{P_{i}^{k-1} P_{i}^{k}}$ at the current object position concerning previous trajectories, and convert it into a unit vector $\mathbf{\hat{a}}_{i}$.
If the angle between $\mathbf{\hat{a}}_{i}$ and $\mathbf{\hat{d}}_{i}$, denoted as $\theta_i$, is not greater than a predefined threshold $\tau_\theta$, we define the head direction of the $i$-th object as a weighted sum $\mathbf{V}_{\mathbf{h},i} = w_1\mathbf{\hat{a}}_{i} + w_2\mathbf{\hat{d}}_{i}$, where $w_1$ and $w_2$ present the weights assigned to the unit vectors.
This approach yields a more accurate and reliable $\mathbf{\hat{V}}_{\mathbf{h},i}$.

\subsubsection{\textbf{Prediction Vector}}
To derive the prediction vector $\mathbf{V}_{\mathbf{p},i}$, we assume that the trajectories maintain consistent momentum within the time interval $\Delta T$.
We select three points $P_{i,m} = \left(x_{i}^{k - mn\Delta T}, y_{i}^{k - mn\Delta T}, z_{i}^{k - mn\Delta T}\right)$, where $m \in \left\{0, 1, 2\right\}$ and $n$ present the frame rate.
We calculate the unit vectors between $P_{i,0}$ and $P_{i,1}$ as $\mathbf{\hat{v}}{i,01}$ and between $P{i,1}$ and $P_{i,2}$ as $\mathbf{\hat{v}}{i,12}$.
The prediction vector is subsequently computed as $\mathbf{V}_{\mathbf{p},i} = w_3 \mathbf{\hat{v}}_{i,01} + w_4 \mathbf{\hat{v}}_{i,12}$, with $w_3$ and $w_4$ denoting the weights assigned to each vector.
Finally, the prediction vector is normalized to yield $\mathbf{\hat{V}}_{\mathbf{p},i}$.

\begin{figure}
     \centering

    \vspace{3pt}

     \begin{subfigure}[thpb]{0.18\textwidth}
         \centering
         \includegraphics[width=\textwidth]{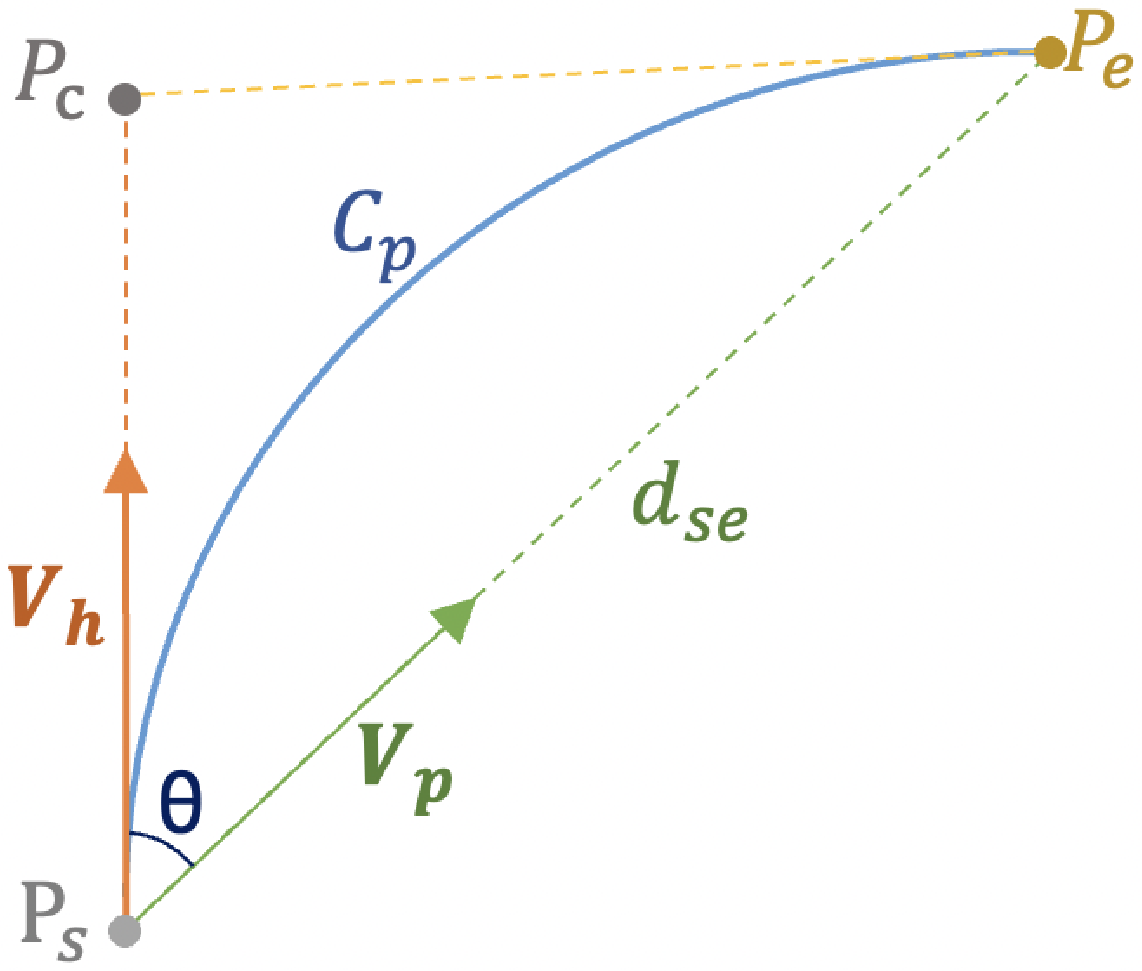}
         \caption{Trajectory Model.}
         \label{fig_bazier}
     \end{subfigure}
     \hfill
     \begin{subfigure}[thpb]{0.25\textwidth}
         \centering
         \includegraphics[width=\textwidth]{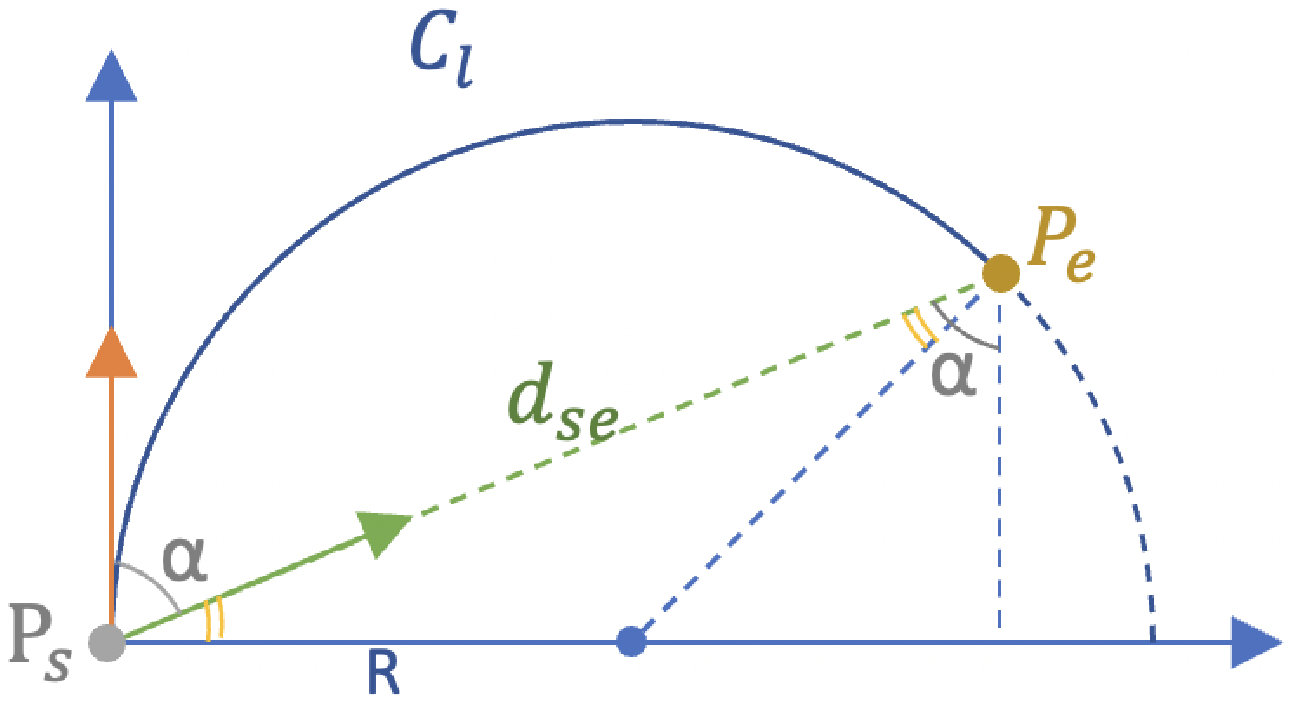}
         \caption{The method to calculate $d_{se}$}
         \label{direction}
     \end{subfigure}
     \hfill
     \caption{(a) The model of trajectory prediction. $P_s$ is the moving object position in the current frame, while $P_e$ is the predicted position. The curve $C_p$ and line $d_{se}$ are the trajectory and distance between them respectively. The unit vector $\mathbf{V_h}$ is the head direction of the object, and $\mathbf{V_p}$ is the prediction direction. (b) The method to calculate $d_{se}$.}
\end{figure}

\subsubsection{\textbf{Trajectory Distance}}
We first preserve the smoothing of the trajectories by following the process detailed later in section \ref{subsec: trajectories updating}.
Utilizing these smoothed trajectories, we proceed to calculate the corresponding smoothed instantaneous velocity of each object in each frame as well as their velocity derivatives.
Subsequently, for each frame, we derive the smoothed instantaneous velocity $v_{i}$ from the past $h$ frames:
\begin{equation}
v_{i} = \frac{\gamma}{h} \sum_{j=k-h}^{k}  \frac{|P_i^{j+1}  - P^j_i|}{\Delta T +\epsilon}
\label{eq_1}
\end{equation}
where $\gamma$ is the smoothing factor, $\Delta T$ denotes the time interval as $\Delta T = t^{j+1} - t^j$, and $\epsilon$ serves as a small constant to prevent instability.
Consequently, the length of the curve $C_{p,i}$ can be calculated as $|C_{p,i}| = v_i \Delta T$.
For each frame, we then obtain a set of lengths corresponding to the predicted curves $\left\{|C_{N}|, \ldots, |C_{M}|\right\}$ for the objects $\left\{O_N, \ldots, O_M\right\}$.

Furthermore, we introduce two additional assumptions. Firstly, as previously mentioned, we assume that trajectories remain constant along the $z$ axis.
Second, if the angle $\alpha_i = \angle (\mathbf{V}_{\mathbf{h},i}, \mathbf{V}_{\mathbf{p},i})$ is less than the threshold $\tau_\alpha$, we consider that the length error $e({|C_{p,i}|,d_{se,i}}) = \left||C_{p,i}| - d_{se,i}\right|$ is neglected. This allows us to approximate the trajectory length $d_{se,i}$ as:
\begin{equation}
    d_{se,i} \approx |C_{p,i}|
    \label{eq_2}
\end{equation}
Conversely, we assume that the error $e({|C_{p,i}|,|C_{l,i}|}) = \left||C_{p,i}| - |C_{l}|\right|$ can be neglected. We then achieve $|C_{l,i}| \approx |C_{p,i}|$, where $C_{l,i}$ corresponds to a segment of the circle shown in Fig. \ref{direction} as $C_{l,i} = \overset{\LARGE\frown} {P_{s,i}P_{e,i}}$.
Given the angle $\alpha_i$ and the length of $C_{l,i}$, we can calculate $d_{se,i}$ using the equation:
\begin{equation}
    d_{se,i} = R_i\sqrt{(cos(\pi - 2\alpha_i) + 1)^2 + sin^2(\pi - 2\alpha_i)}
    \label{eq_3}
\end{equation}
where $R_i = |C_{l,i}| / (2\alpha_i)$.

After obtaining head direction vector $\mathbf{\hat{V}}_{\mathbf{h},i}$, prediction vector $\mathbf{\hat{V}}_{\mathbf{p},i}$, and trajectory distance $d_{se,i}$ for each object $O_i$, we proceed to compute the distance $d_{sc,i}$ between $P_{s,i}$ and $P_{c,i}$, which we set to half of $|C_{p,i}|$.
Thus, the trajectory prediction $C_{p,i}$ for object $O_i$ can be derived as follows:
\begin{equation}
    \begin{aligned}
    P_{s,i} &= P_{i}^k \\
    P_{c,i} &= P_{s,i} + d_{sc,i}\cdot \mathbf{\hat{V}}_{\mathbf{h},i}	\\
    P_{e,i} &= P_{s,i} + d_{se,i}\cdot \mathbf{\hat{V}}_{\mathbf{p},i} 	\\
    C_{p,i} &= \text{Bezier} (P_{s,i}, P_{c,i}, P_{e,i})
    \end{aligned}
    \label{eq_4}
\end{equation}
As a result, for each frame, we obtain a series of predicted trajectories $\left\{C_{N}, \ldots, C_{M} \right\}$ at the current frame.

\subsection{Moving Objects Trajectories Updating}
\label{subsec: trajectories updating}

The accuracy of previous trajectories is pivotal in predicting future trajectories of moving objects within the scene.
The trajectories of moving 3D objects should exhibit consistency and maintain smooth curves within 3D scenes.
One challenge arises from the fact that neural-network generated 3D bounding boxes are created frame by frame. This leads to a misalignment between the trajectory consisting of discrete object central points and the continuous camera motion.
Consequently, the motion of these moving objects appears erratic within the 3D scenes.
This motivates us to establish motion priors, aiming to optimize and smoothen the trajectories.

\subsubsection{\textbf{3D Trajectory Motion Priors}}
\label{TrajectoriesBA}
% 论文原文：数学上，最小动能先验鼓励恒定速度运动，最小力先验促进恒定加速度运动，而最小作用先验则有利于抛物线运动。虽然这些先验在力在整个操作时间内任意施加于系统的主动系统中并不成立，但我们推测机械和生物系统施加的累积力是稀疏的且持续时间很短，真实轨迹可以由最小化我们运动先验所定义的代价的路径来近似。在3D轨迹中的任何局部误差，无论是沿轨迹的点的估计不准确，还是不同相机观察到的点之间的错误时间排序，都会产生更高的运动先验成本。
It has been proved that, within small time intervals, the accurate trajectories can be determined by identifying the path by minimizing costs associated with predefined priors, as demonstrated in \cite{vo2016spatiotemporal}.
Therefore, we leverage the least kinetic motion prior cost to optimize prior trajectories.
For a moving object denoted as $O_i$ within the time interval $\Delta T$, the least kinetic motion prior cost can be expressed as $Q_i = \frac{1}{2} r_i \overline{v_i}^2 \Delta T$, where $r_i$ represents the weight associated with object $O_i$, and $\overline{v_i}$ denotes the velocity of object $O_i$ during the time interval $\Delta T$.
Hence, at the current frame $F_k$ we define the cost of previous trajectories as:
\begin{equation}
Q_{i, k} =  \sum_{j=k-h}^{k} \frac{1}{2} r_i  \left(\frac{ |P_i^{j+1}  - P^j_i |}{t^{j+1} - t^j +\epsilon }\right)   ^2 \left(t^{j+1}-t^j\right)
\label{eq_update1}
\end{equation}

Next, we obtain the Jacobian matrix of the cost function with respect to points $P_i^{j+1}$ and $P_i^{j}$ as:
\begin{small}
\begin{equation}
    \frac{\partial Q_{i, k}}{ \partial  P_i^{j+1} } = \frac{r_i }{ |P_i^{j+1}  - P^j_i |\Delta T} \left [ x_i^{j+1} - x_i^{j} ,  \; \;y_i^{j+1} - y_i^{j} ,\; \;z_i^{j+1} - z_i^{j} \right ]
\label{eq_update2.1}
\end{equation}
\begin{equation}
    \frac{\partial Q_{i, k}}{ \partial  P^j_i } = \frac{-r_i }{|P_i^{j+1}  - P^j_i |\Delta T} \left [ x_i^{j+1} - x_i^{j} ,  \; \;y_i^{j+1} - y_i^{j} ,\; \;z_i^{j+1} - z_i^{j} \right ]
\label{eq_update2.2}
\end{equation}
\end{small}
With the Jacobian matrix,  we use Levenberg-Marquardt \cite{more2006levenberg} to refine the trajectories between frame $F_{k - b}$ and $F_k$. Thus we get the optimized trajectories $\left\{S_i\right\}$ along with the updated 3D points $\{P_i^j \}_{j=k-h}^k$, where $i \in \{N, ...,  M\}$.

\subsubsection{\textbf{Initialization}}
\label{Initialization}
As mentioned earlier, the acquisition of raw moving object trajectories, denoted as $\left\{S_N, \ldots, S_M\right\}$, from each frame generated by the neural-network can introduce significant noise, potentially leading to pronounced jittering in trajectory estimation.
To mitigate this issue, a preliminary step is undertaken before applying the trajectory update method described above.
This initial step involves the use of Kalman Filters (KF) \cite{welch1995introduction} to initialize the generated 3D points.
The Kalman Filters are employed to estimate the temporal state of each object, with one linear filter dedicated to each object and a total of 18 states considered.
Additionally, when the length of the trajectory falls below a predefined threshold, denoted as $\tau_{KF}$, instead of directly updating the trajectories, we opt for the application of n-point moving average (NPMA) filters on top of the Kalman Filters.
This approach helps mitigate the adverse effects of the initial few 3D points in the trajectory.

\subsubsection{\textbf{Trajectory Consistency Check}}
\label{consistency}
The object detection system, which relies on neural network methods, may not always be reliable and can encounter intermittent tracking failures.
This often leads to inconsistent and inaccurate object tracking.
Given this context, we check the consistency of object positions with previous trajectories and apply a Gaussian filter $G(P) \sim \mathcal{N} (\mu, \sigma^2)$ to smooth the noisy position data $P$.
When a new object is detected in frame $F_k$, we initialize a new trajectory and merge it to any previously detected trajectories if the trajectories $\left\{C_{p,N}, \ldots, C_{p,M} \right\}$ predicted in the previous frame $F_{k - 1}$ align with the positions within the probability distribution $G(P)$.
Otherwise, we update the existing trajectory.
In cases where two trajectories cross, we choose the one closest to the center of the distribution.

\section{Experiments}

\subsection{Experimental Setup}
\label{setup}

\begin{figure*}[tp]
    \centering

    \vspace{8pt}

    \begin{subfigure}[b]{0.31\textwidth}
        \centering
        \includegraphics[width=\textwidth]{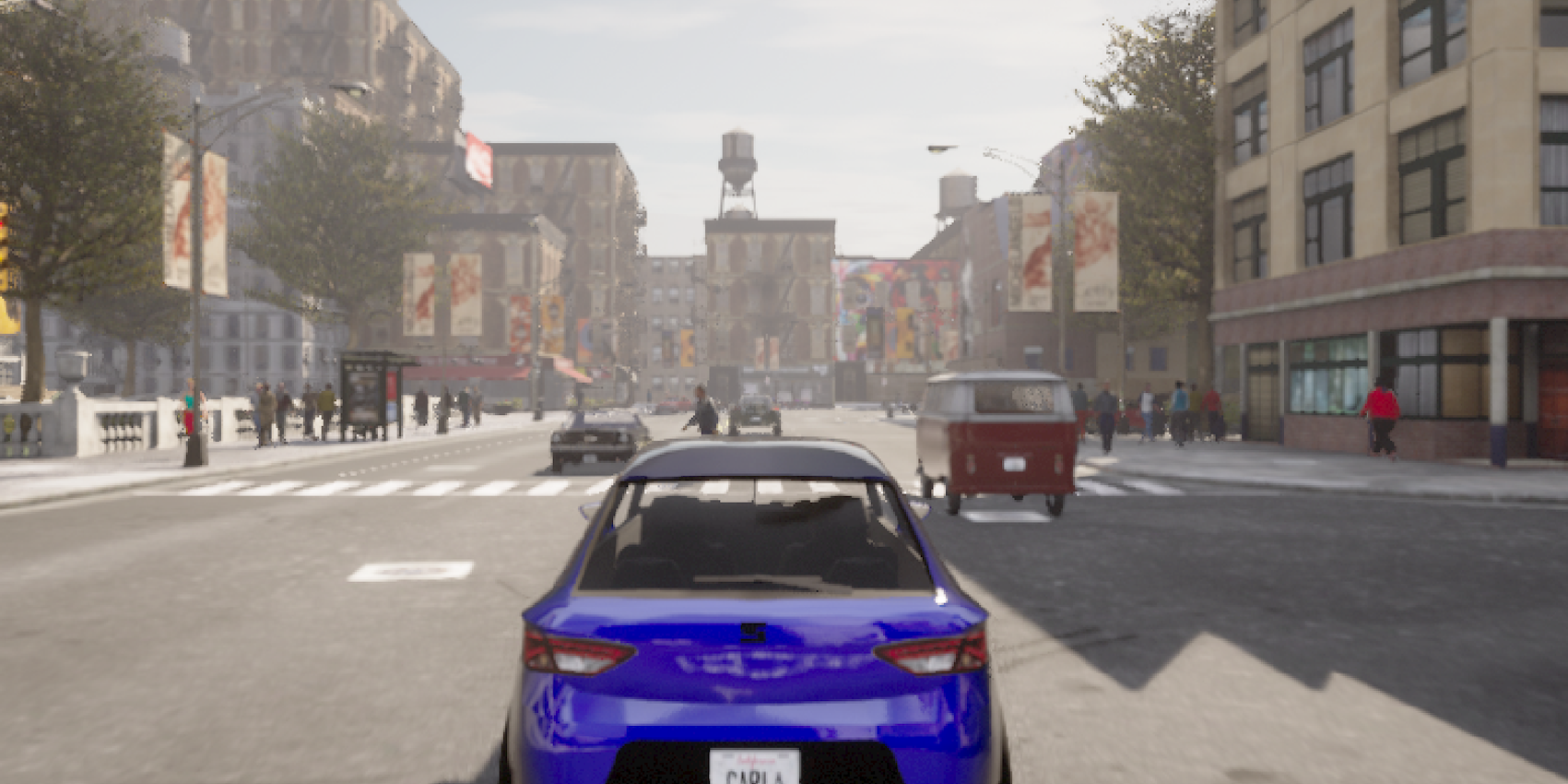}
        \caption{CARLA simulator}
        \label{carla_env}
    \end{subfigure}
    \begin{subfigure}[b]{0.31\textwidth}
        \centering
        \includegraphics[width=\textwidth]{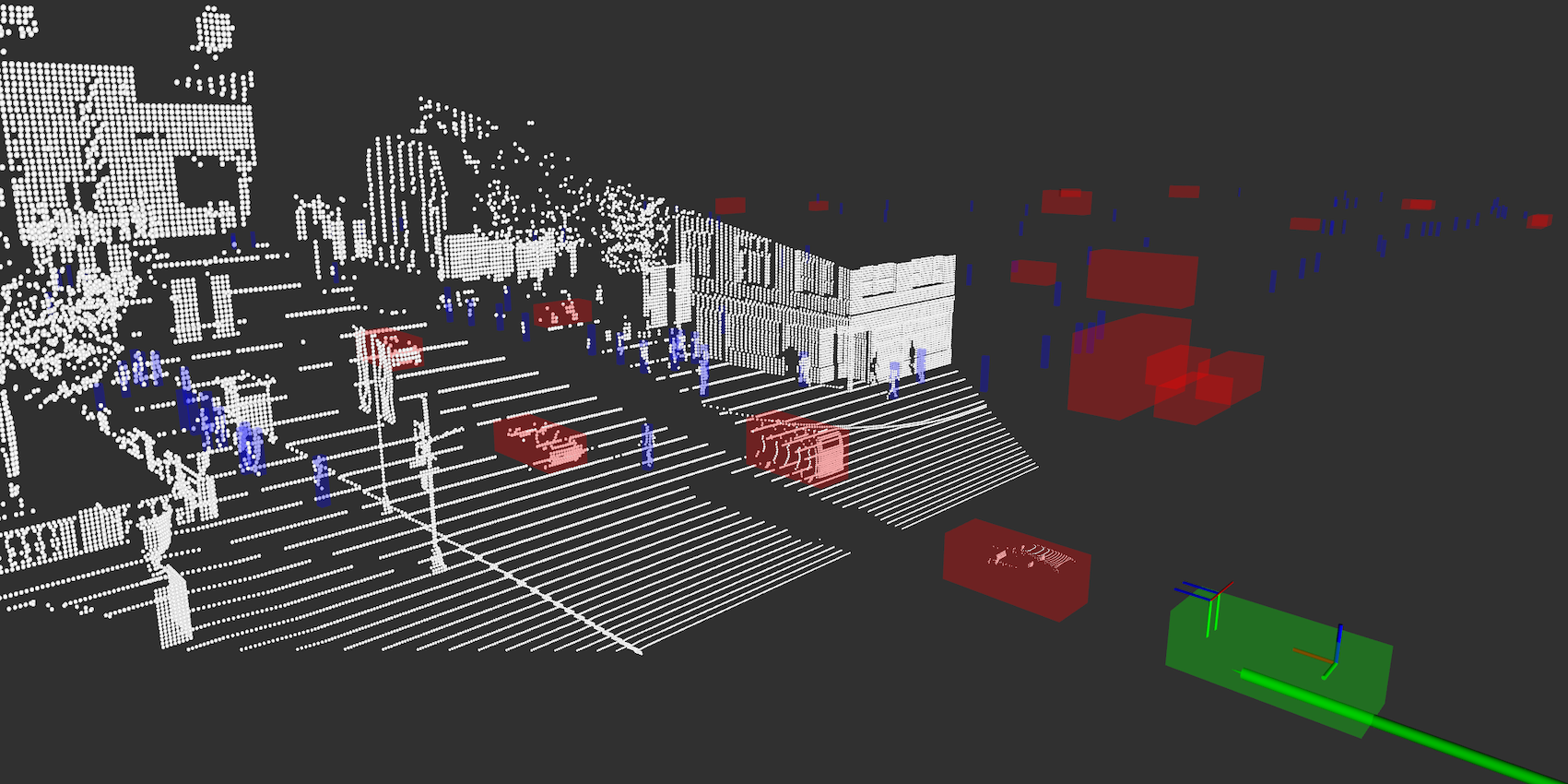}
        \caption{Ground truth visualization}
        \label{carla_rviz}
    \end{subfigure}
    \begin{subfigure}[b]{0.31\textwidth}
        \centering
        \includegraphics[width=\textwidth]{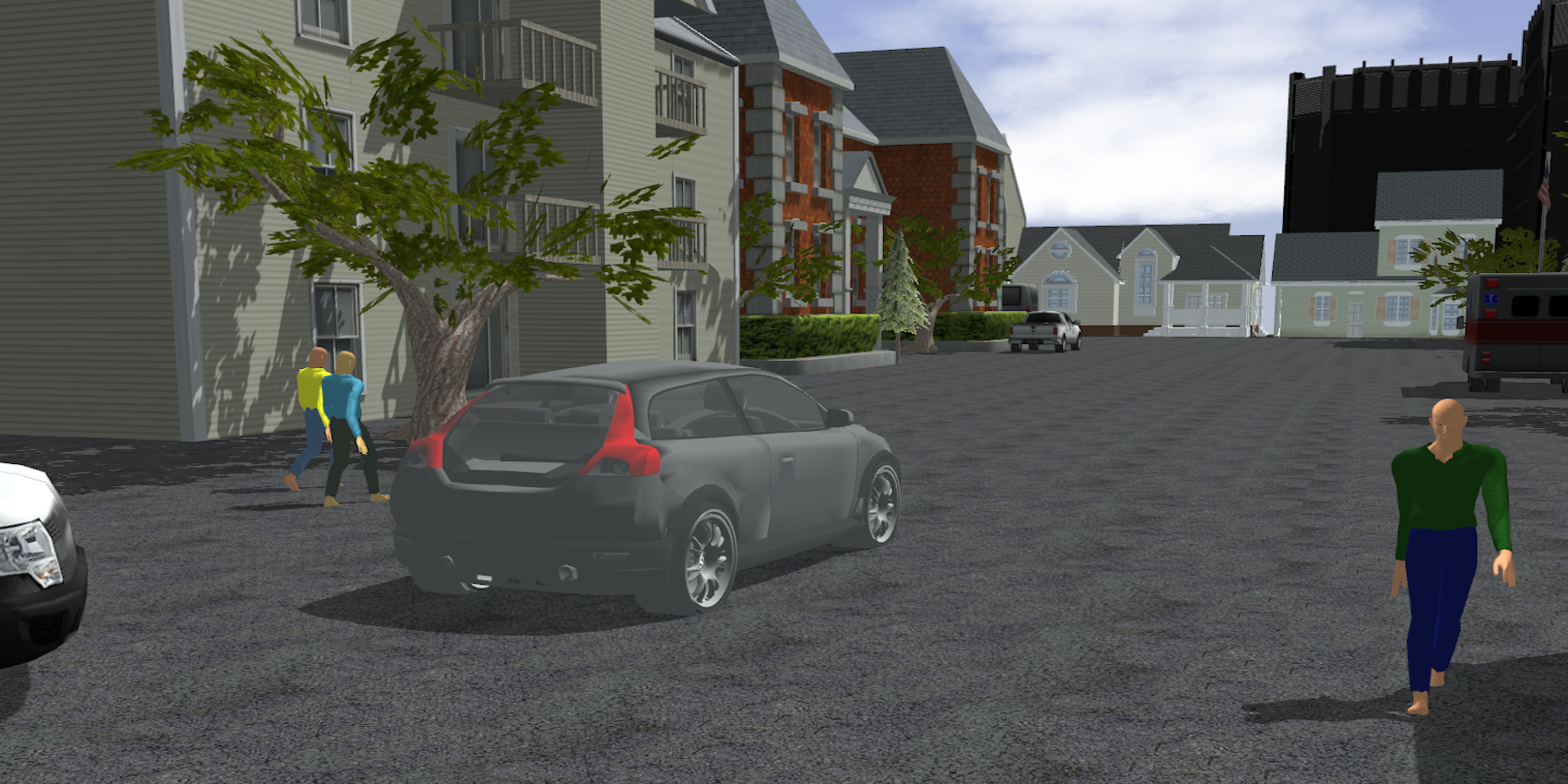}
        \caption{GAZEBO simulator}
        \label{gazebo_env}
    \end{subfigure}
    \caption{(a) The CARLA simulator spawns a large number of pedestrians and cars and generates images with ego vehicle view. (b) The object detection and trajectory ground truth are visualized. The green, red and blue boxes denotes the ego vehicle, generated cars and generated pedestrians, respectively. (c) The GAZEBO simulator is also established for evaluation.}
    \label{sim_envs}
\end{figure*}

\textbf{Simulation and real-world datasets.}
We consider two categories of benchmark datasets: simulated outdoor environments implemented by CARLA \cite{dosovitskiy2017carla} and GAZEBO \cite{koenig2004design} integrated with the ROS \cite{quigley2009ros} system, and the real-world KITTI datasets \cite{geiger2013vision}.
In CARLA, we designed diverse scenarios featuring over 400 moving pedestrians and vehicles, comprising nearly 10,000 frames.
These scenarios include highly dynamic sequences labelled as CARLA 01$\sim$05 and a closed loop sequence called CARLA Easy Loop (CARLA EZL).
An example of our CARLA environments is depicted in Fig. \ref{carla_env}.
We also utilized the GAZEBO environment that simulated the KITTI stereo camera setup.
This environment also features a loop in the scene, designated as GAZEBO Loop (GAZEBO L).
Furthermore, we incorporated outdoor sequences from the MOTChallenge dataset \cite{Voigtlaender2019CVPR} \cite{Luiten2020IJCV} within the KITTI dataset for multi-object tracking and evaluation.
Both real-world and simulation datasets provide images at a 10 fps frame rate and ground truth for camera poses and moving object trajectories.
All the experiments are conducted on a computer with AMD 5600X CPU, 2080ti GPU, with 12GB memory.

\textbf{Baselines.}
We build our visual SLAM system upon ORB-SLAM2 and use the open-source system QD-3DT \cite{hu2022monocular} for neural-network generated 2D and 3D bounding boxes application.
We also compare our system with ORB-SLAM3\cite{mur2017orb}, an upgraded version of ORB-SLAM2, and discover their limitation when observing large moving objects from the scene.
Additionally, we evaluate our system against DynaSLAM \cite{bescos2018dynaslam}, an open-source dynamic vision-based SLAM system.
The calculated evaluation metrics include Root Mean Square Error (RMSE), Mean, and Standard Deviation (SD).

In trajectory prediction evaluation, we adopt a 3-second prediction horizon based on the Control-Aware Prediction Objectives (CAPOs) \cite{mcallister2022control} findings for effective autonomous driving collision avoidance.
Due to limited prior research exploring the integration of object trajectory prediction within SLAM systems, we establish the baselines by predicting the trajectories for a 3-second interval and fitting a 3D polynomial curve.
We fit the curve to 4 evenly spaced points on the previous trajectory (with 5 frames in between) and determine the endpoints $d_{se}$ distance away from the start points.
The Average Displacement Error (ADE) and Final Displacement Error (FDE) are calculated on both the KITTI and simulated datasets.

For trajectory evaluation, we employ QD-3DT \cite{hu2022monocular} as our baseline and calculate RMSE of the trajectories. Furthermore, leveraging the trajectory prediction, we assess the improvement in Precision (P), Recall, and F1-score for moving object detection when considering trajectory consistency.

\textbf{Implement details.}
The threshold $\tau_\theta$ for head direction vector $\mathbf{V}_{\mathbf{h},i}$ is set to $\pi/2$, and $\mathbf{V}_{\mathbf{h},i}= 0.9 \mathbf{\hat{a}}_{i} + 0.1 \mathbf{\hat{d}}_i$.
For the prediction vector $\mathbf{V}_{\mathbf{p},i}$, we set $\mathbf{V}_{\mathbf{h},i} = -2 \mathbf{\hat{v}}_{i,01} + \mathbf{\hat{v}}_{i,12}$.
For the distance threshold, we set $\tau_\alpha = \pi/18$.
The Kalman Filter's weights are set to $w_k = 0.5$ and $w_n = 0.5$.
Moreover, we set the $\tau_{KF} = 7$ in the Algorithm \ref{alg:overview} and $h = 20$.
In Equation \ref{eq_update1}, we assign $r_{pedestrians} = 80$ as the average weight for pedestrians and $r_{car} = 1900$ for cars.

\begin{figure*}[tp]
    \centering

    \vspace{5pt}

    \begin{subfigure}{.24\textwidth}
        \includegraphics[width=\linewidth]{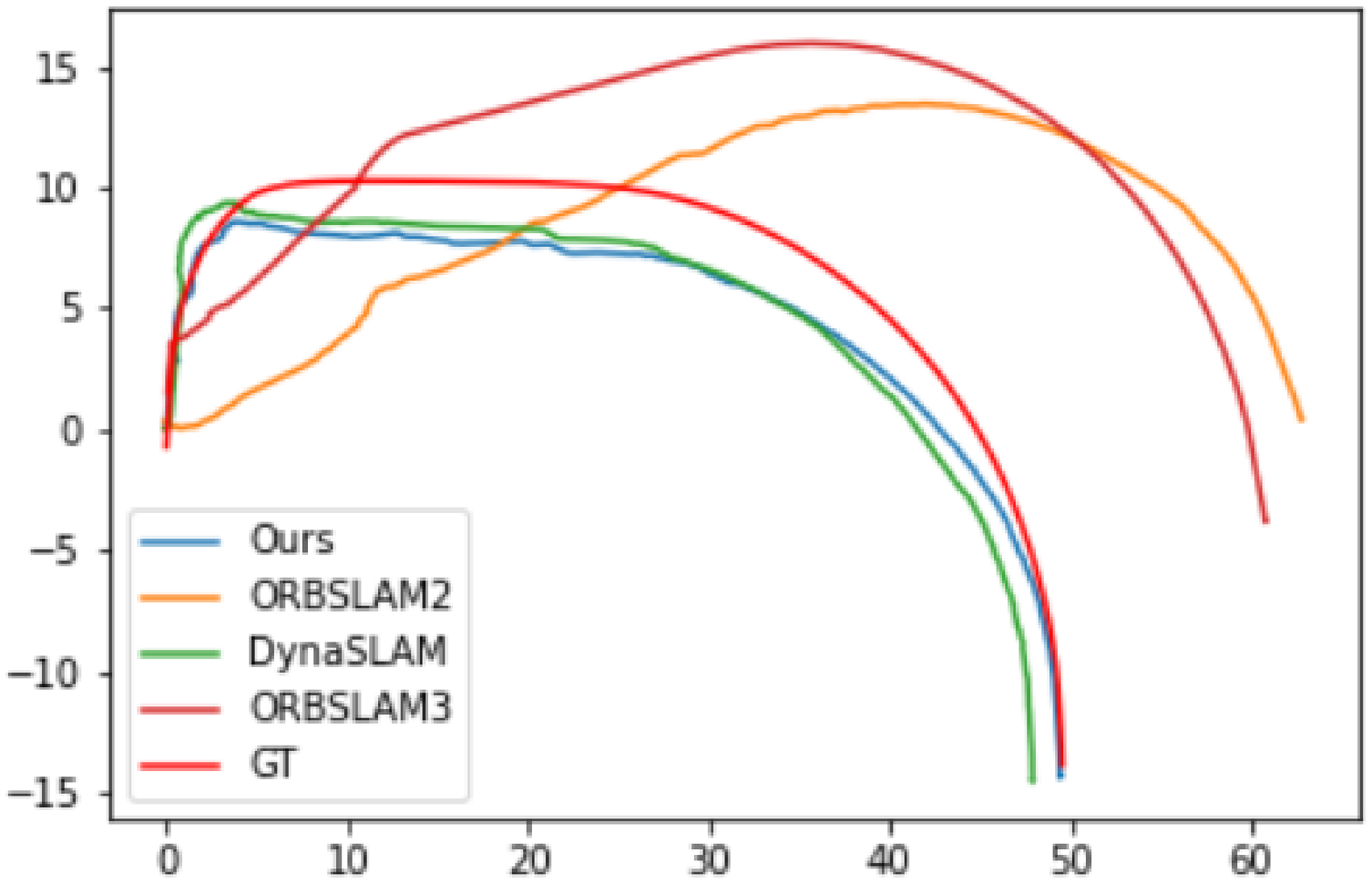}
        \caption{CARLA 01}\label{fig:Label10}
    \end{subfigure}
        \hfill
    \begin{subfigure}{.24\textwidth}
        \includegraphics[width=\linewidth]{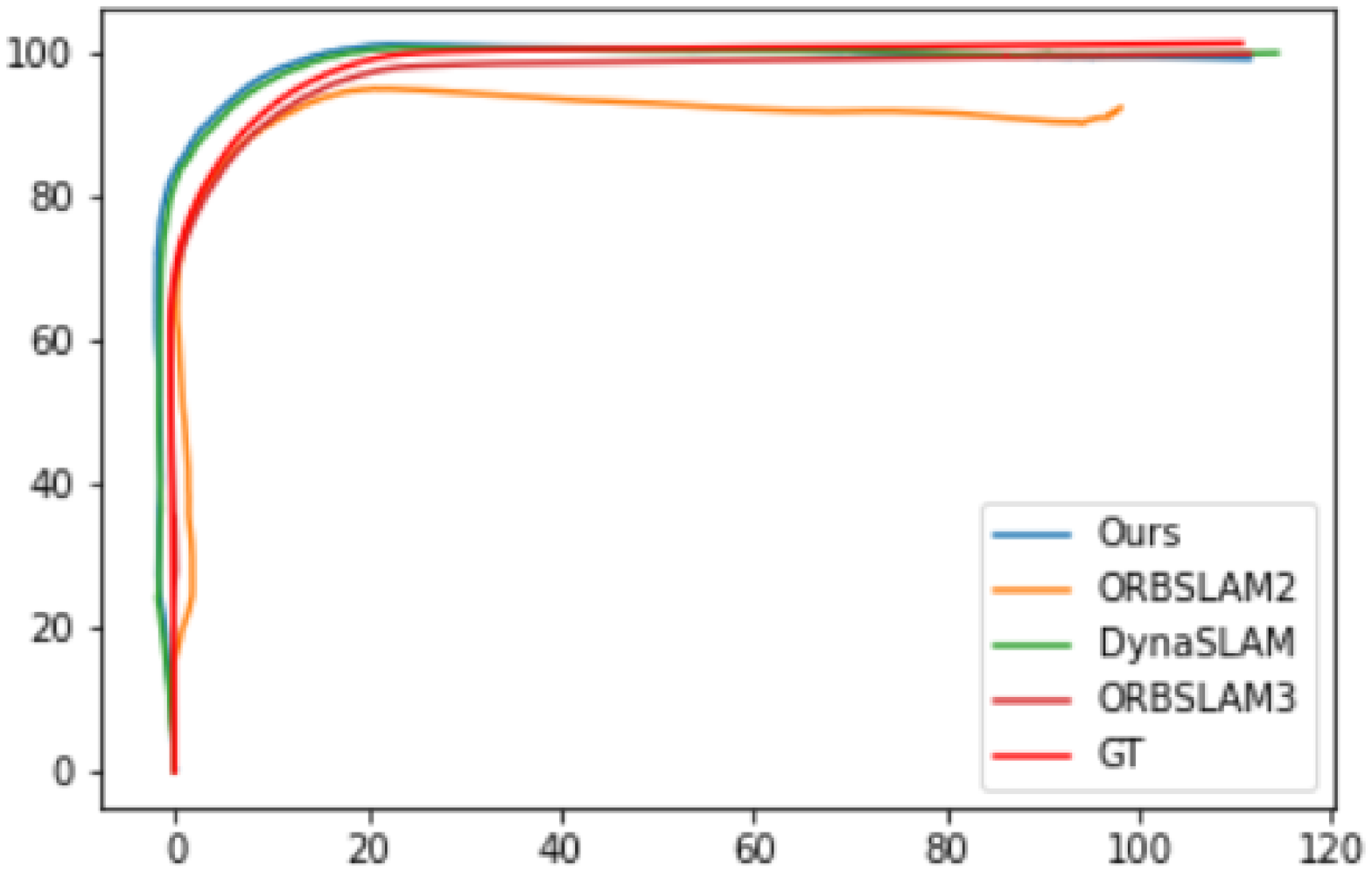}
        \caption{CARLA 03}\label{fig:Label11}
    \end{subfigure}
        \hfill
    \begin{subfigure}{.24\textwidth}
        \includegraphics[width=\linewidth]{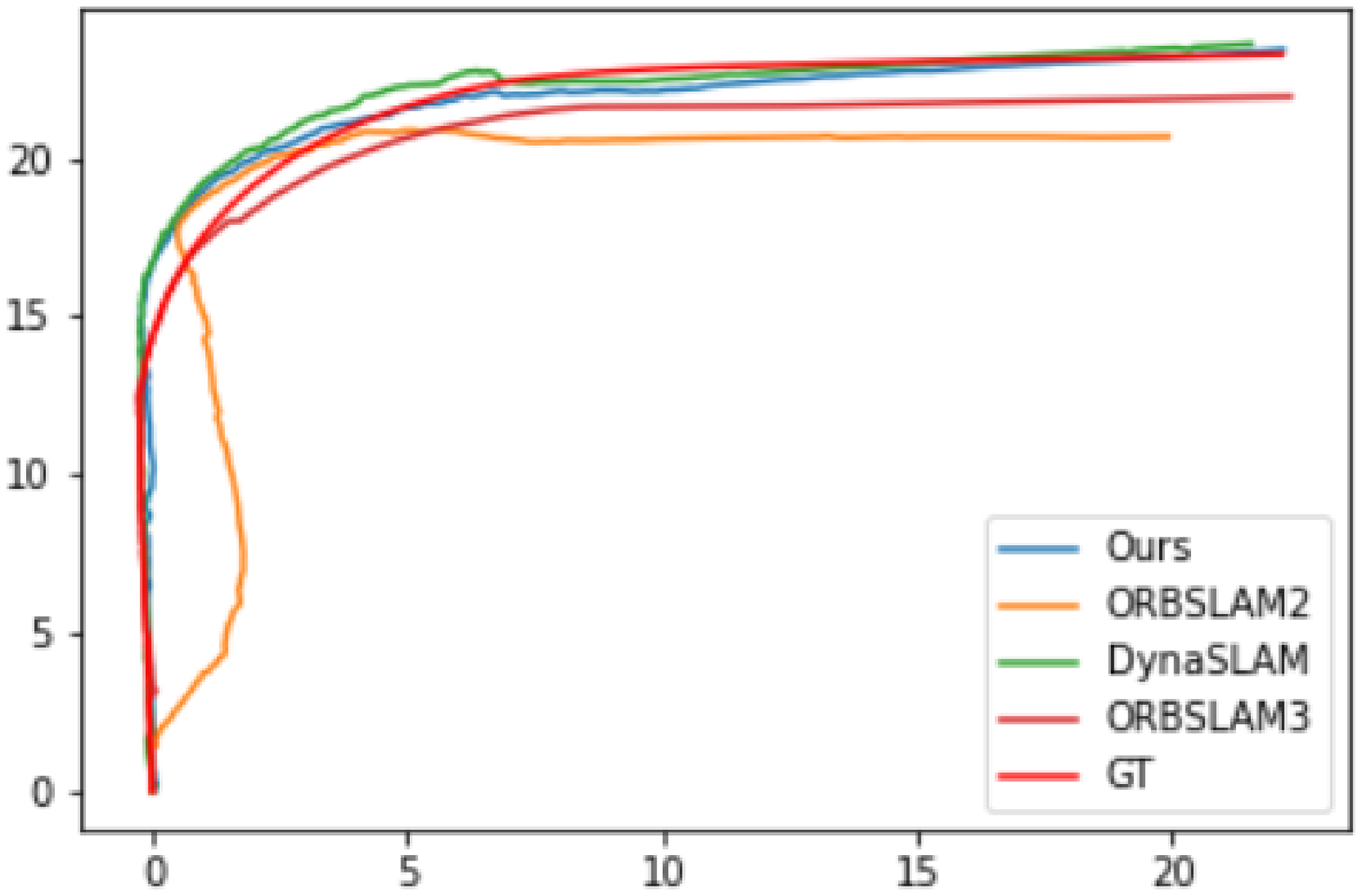}
        \caption{CARLA 05}\label{fig:Label12}
    \end{subfigure}
    \hfill
    \begin{subfigure}{.24\textwidth}
        \includegraphics[width=\linewidth]{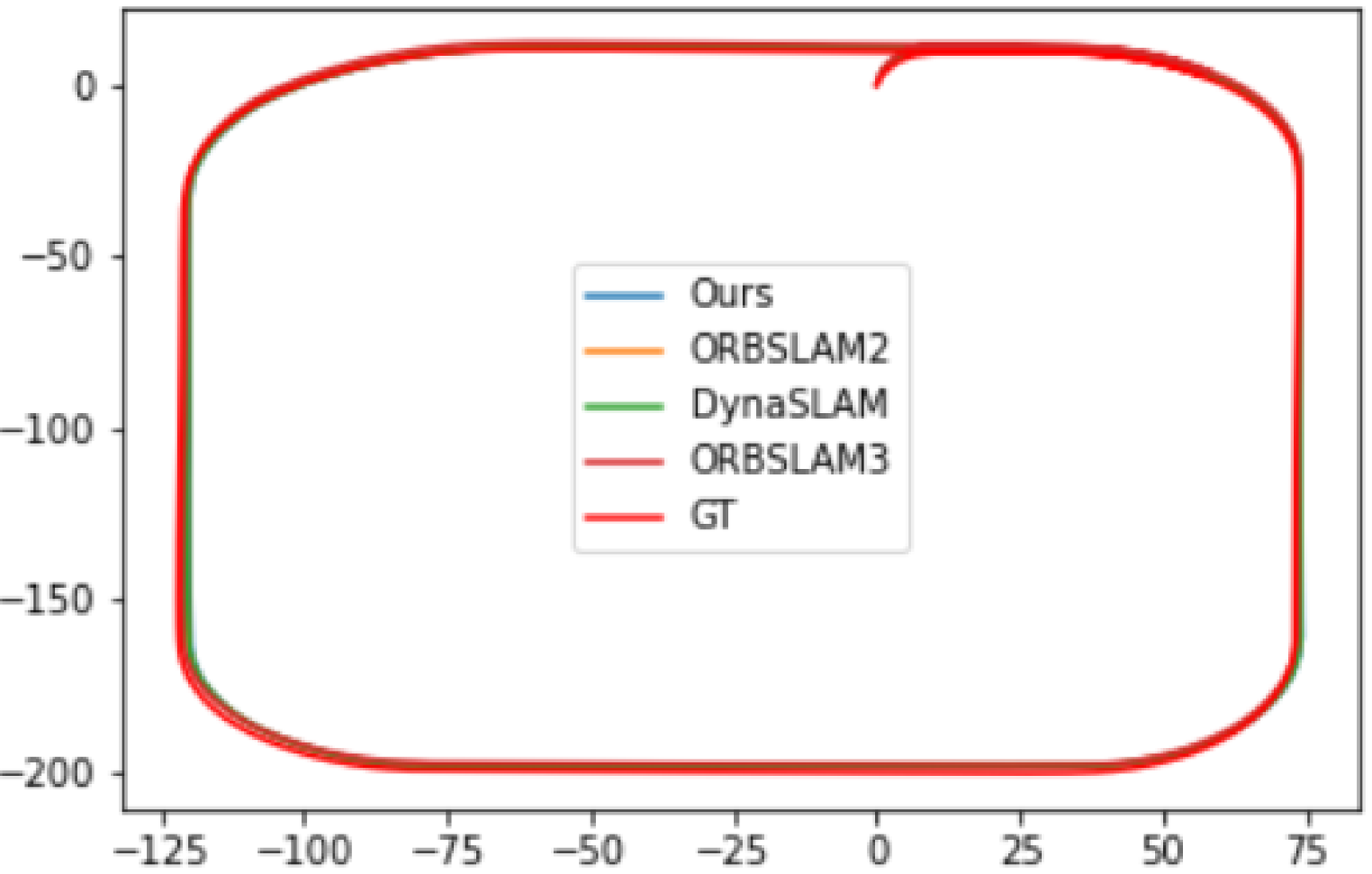}
        \caption{CARLA EZL}\label{fig:Label12}
    \end{subfigure}
    \caption{Examples of camera pose on simulator sequences. }
    \label{camerapose_fig}
\end{figure*}

\begin{table*}[tp]
\caption{ Camera pose of SLAM systems (RMSE, Mean and SD). }
\begin{center}
\begin{tabular}{c|ccc|ccc|ccc|ccc}
\hline
\hline
\multirow{2}*{}&\multicolumn{3}{c|}{Ours } & \multicolumn{3}{c|}{DynaSLAM } & \multicolumn{3}{c|}{ORBSLAM2 } & \multicolumn{3}{c}{ORBSLAM3 }\\
            & RMSE  $\downarrow$  & Mean  $\downarrow$ & SD  $\downarrow$ &RMSE  $\downarrow$ &Mean  $\downarrow$ & SD  $\downarrow$ & RMSE  $\downarrow$  & Mean  $\downarrow$ & SD  $\downarrow$ &RMSE  $\downarrow$ & Mean $\downarrow$ & SD $\downarrow$\\

\hline
CARLA01     &\textbf{4.791} &\textbf{4.204} &\textbf{2.656} &\uline{5.041} &\uline{4.387} &\uline{2.762} &14.679 & 11.059 &7.299 &11.959 &9.054 & 5.469\\
CARLA02     & \uline{5.062} &\uline{4.902} &\uline{2.491} &5.365 &5.174 &2.655 &6.391  & 7.254  &2.179 &\textbf{4.445}  &\textbf{2.426} & \textbf{2.188}\\
CARLA03     & \uline{5.804}         &\uline{4.902} &\uline{3.410} &5.972 &5.093 &3.687 &9.606  &8.973   &3.431 &\textbf{4.488}  &\textbf{2.933} & \textbf{2.845}\\
CARLA04     & 3.145        &3.059 &0.728 &\uline{2.985} &\uline{2.902} &\uline{0.648} &4.573  &4.248   &1.691 & \textbf{2.893} &\textbf{2.348} & \textbf{0.835}\\
CARLA05     &\textbf{1.390} &\textbf{1.048} &\textbf{0.916} &1.573 &1.217 &1.000 &1.972  &1.380   &1.412 & \uline{1.396} &\uline{1.057} & \uline{0.940}\\
CARLA EZL   &\uline{3.663} &\textbf{1.805} &\textbf{0.741} &3.689 &\uline{2.033} &\uline{0.794} & 3.844 &1.947   &0.906 & \textbf{3.625} & 2.634& 1.029\\
GAZEBO L & \textbf{0.477}   &\textbf{0.442} &\textbf{0.177} &\uline{0.480} &\textbf{0.442} &\textbf{0.177} & 0.738 &0.646   &0.343 & 0.482     & 0.444    & 0.181  \\
\hline
\hline
\end{tabular}
\end{center}
\label{camerapose_table}
\end{table*}

\begin{table*}[tp]
\caption{ Result of trajectory Prediction (ADE, FDE).}
\label{camerapose}
\begin{center}
\begin{tabular}{c|cc|cc|cc|cc|cc}
\hline
\hline
\multirow{2}*{}&\multicolumn{2}{c|}{Baseline }   & \multicolumn{2}{c|}{Ours - DM } & \multicolumn{2}{c|}{Ours - HD } & \multicolumn{2}{c|}{Ours - SV }& \multicolumn{2}{c}{Ours}\\
            & ADE  $\downarrow$  &FDE  $\downarrow$ &ADE  $\downarrow$ & FDE  $\downarrow$ & ADE  $\downarrow$  & FDE  $\downarrow$ & ADE  $\downarrow$ &FDE  $\downarrow$ & ADE $\downarrow$ & FDE $\downarrow$\\
          % Baseline         capos         dm            hd              sv           our
\hline
MOT19       &0.312  &0.586  &\textbf{0.186} &0.423 &0.326 & 0.615 &0.276 &0.699 &\textbf{0.186} & \textbf{0.391}\\
MOT20       & 0.732 &1.231    &0.435 &0.741 &1.279  & 1.429  &0.853 &2.177  &\textbf{0.416}& \textbf{0.701}\\
GAZEBO L    & 0.403 &0.702   &\textbf{0.276} &0.558 &1.593 & 1.769 & 0.486 & 1.208   & \textbf{0.276} & \textbf{0.533}  \\
CARLA01     &1.336 &1.536    &1.065 &1.201 & 2.601 &2.968 &1.576 &1.968 &\textbf{1.036} & \textbf{1.167}\\
CARLA02     &1.045 &1.264    &0.864 &\textbf{1.063} &2.068  & 2.631  &1.136 &1.769  &\textbf{0.831} & 1.103\\
CARLA03     & 0.372 &0.603    &0.196 &0.438 &0.486  &0.772   &0.301 &0.723  &\textbf{0.190} & \textbf{0.429}\\
CARLA04     & 0.698 &0.862    &0.456 &0.791 &1.367  &1.582   &1.191 & 1.946 &\textbf{0.436} & \textbf{0.787}\\
CARLA05     &0.599 &0.841   &0.321 &0.601 &0.936  &1.280   &0.983 & 1.096 &\textbf{0.316} & \textbf{0.579}\\
CARLA EZL   &0.863 &1.027    &0.749 &0.923 & 1.563 &1.747   &1.639 & 2.001 & \textbf{0.721}& \textbf{0.893}\\
CARLA AVG   &0.829& 1.022  &\textbf{0.610} & 0.861&1.555  &1.890  & 1.131  & 1.585  & 0.614&\textbf{0.849 }\\
\hline
\hline
\end{tabular}
\end{center}
\label{predictions_table}
\end{table*}

\begin{figure}[tp]
    \begin{subfigure}{0.1\textwidth}
        \centering
        \includegraphics[width=\textwidth]{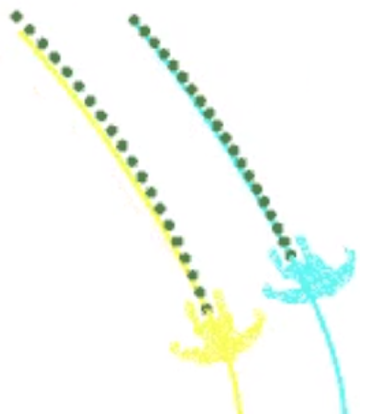}
        \caption{}
        \label{prediction_human}
    \end{subfigure}
    \hfill
    \begin{subfigure}{0.35\textwidth}
    \centering
        \includegraphics[width=\textwidth]{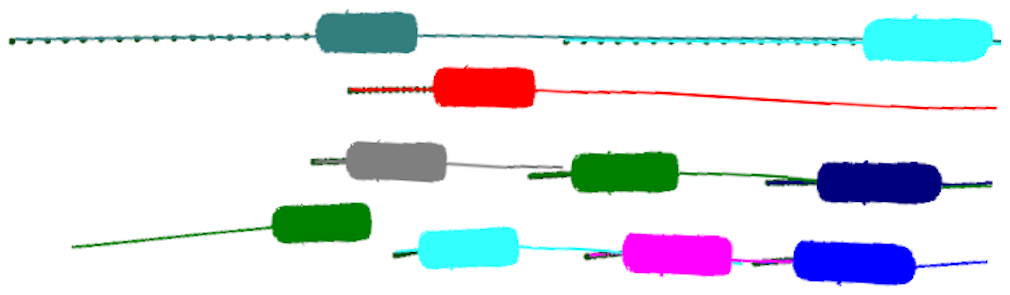}
        \caption{}
        \label{prediction_cars}
    \end{subfigure}
    \hfill
    \caption{Results of trajectory prediction. The green dashes are the predicted trajectories, while the ground truth is the lines with the same colors as the trajectories.}
    \label{predictions_fig}
\end{figure}

\subsection{Camera Pose}

Fig. \ref{camerapose_fig} provides visual comparisons of camera pose estimation results between our proposed method, ORB-SLAM2 \cite{mur2015orb}, DynaSLAM \cite{bescos2018dynaslam}, ORB-SLAM3 \cite{mur2017orb}, and the ground truth.
Notably, our system exhibits reduced drift, and the incorporation of loop closure detection contributes to a decrease in estimation errors.
The evaluation results are shown in Table \ref{camerapose_table}.
It is worth noting that the MOT KITTI dataset mentioned in Section \ref{setup} lacks ground truth odometry information, making comparisons feasible only on our simulator dataset.
ORBSLAM3 performs well with an additional IMU sensor in most sequences, except for CARLA01.
In CARLA01, both our method and DynaSLAM outperform other baselines, mainly because of a large number of moving objects at the very beginning.
ORBSLAM3, lacking the ability to recover clues from external moving objects, struggles in this scenario despite the IMU sensor.
Among the exclusively vision-based methods, our approach consistently achieves top or near-top precision and effectively handles distractions caused by abnormal moving objects in the scene.

\subsection{Trajectory Prediction}

The trajectories predicted by our method are displayed in Fig. \ref{predictions_fig} where the green dashed lines represent the predicted positions, and the lines with the same color as the objects indicate the ground truth.
As shown in Fig. \ref{prediction_human}, the predicted future paths of pedestrians closely match the ground truth curve.
Additionally, the predicted vehicle trajectories in Fig. \ref{prediction_cars} also follow the original driving path.
These results demonstrate the robustness of our method and its ability to accurately predict both curved and straight trajectories for objects moving at different speeds.

Our trajectory prediction approach, as seen in Table \ref{predictions_table}, outperforms the baseline method significantly, both in terms of ADE and FDE.
The baseline set the distance $d_{se}$ equal to the curve length $C_p$, similar to Ours-DM.
The effectiveness of our head direction vector-based trajectory modelling is evident when comparing Ours-DM and Ours.
In contrast, compared to Ours-HD, our method demonstrates greater robustness, especially in cases where the head direction from QD-3DT \cite{hu2022monocular} may be less reliable (as shown in Fig. \ref{consisitency_fig}).
Our velocity smoothing procedure proves necessary, as instant velocity often contains noise.
In a similar CARLA simulation environment, the original CAPOs \cite{mcallister2022control} paper reports ADE of predictions over 2m, while our results are improved by an order of magnitude.
Notably, our trajectory prediction relies on geometry and requires no training, taking approximately $2.5^{-7}$ s per frame. These findings underscore the strength and potential of our method for trajectory prediction in various real-world applications.

\begin{table*}[tp]

\vspace{8pt}

\caption{Result of trajectory evaluation with RMSE, Precision, Recall and F1 value. }
\label{RMSE}
\begin{center}
\scalebox{1}{
\begin{tabular}{c|cccc|cccc|cccc}
\hline
\hline
 \multirow{2}*{}&\multicolumn{4}{c|}{QD-3DT } & \multicolumn{4}{c|}{Ours + QD-3DT }& \multicolumn{4}{c}{Increase } \\
            & RMSE $\downarrow$& P $\uparrow$ & Recall$\uparrow$ & F1 $\uparrow$ &  RMSE $\downarrow$& P $\uparrow$ & Recall$\uparrow$ & F1 $\uparrow$ &  RMSE $\uparrow$& P $\uparrow$ & Recall$\uparrow$ & F1 $\uparrow$     \\
\hline
KITTI MOT19  & 0.785 & 0.179& \textbf{0.966} &0.302& \textbf{0.642}& \textbf{0.181}& \textbf{0.996}&\textbf{0.305} & 18.21\%&1.11\%&0\%&0.98\%\\
KITTI MOT20  & 0.582         & 0.578          & \textbf{0.992}          &0.729            & \textbf{0.505}          &\textbf{0.584}      & \textbf{0.992}       & \textbf{0.735}&13.23\%&1.3\%&0\%&0.81\%  \\
CARLA 01-05 &1.086       & 0.389             & \textbf{0.881}& 0.538           & \textbf{0.863}  &\textbf{0.425}& \textbf{0.881}&\textbf{0.573}&20.53\%&8.47\%&0\%&6.11\%\\
CARLA EZL  &0.842        &0.351             &\textbf{0.882}       &0.502             &\textbf{0.669}          & \textbf{0.384}        &0.876         &\textbf{0.504}&20.55\%&8.59\%&-0.68\%&0.39\%\\
GAZEBO L &1.105       &0.161              &\textbf{0.926}         &0.274           & \textbf{0.935}            & \textbf{0.180}        &\textbf{0.926}         &\textbf{0.301}&15.38\%&10.56\%&0\%&8.97\%\\
\hline
\hline
\end{tabular}}
\end{center}
\label{RMSE}
\end{table*}

\subsection{Object Trajectory}
\begin{figure}[tp]
    \centering
    \begin{subfigure}{0.45\textwidth}
        \centering
        \includegraphics[width=\textwidth]{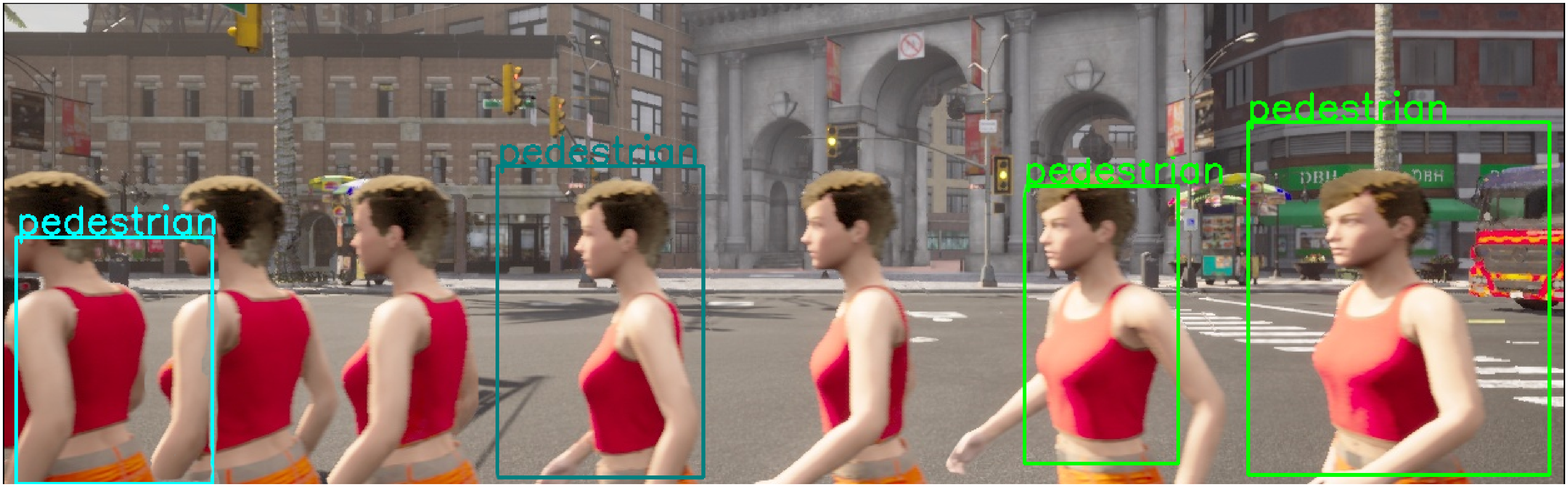}
    \end{subfigure}

    \vspace{3pt}

    \begin{subfigure}{0.45\textwidth}
        \centering
        \includegraphics[width=\textwidth]{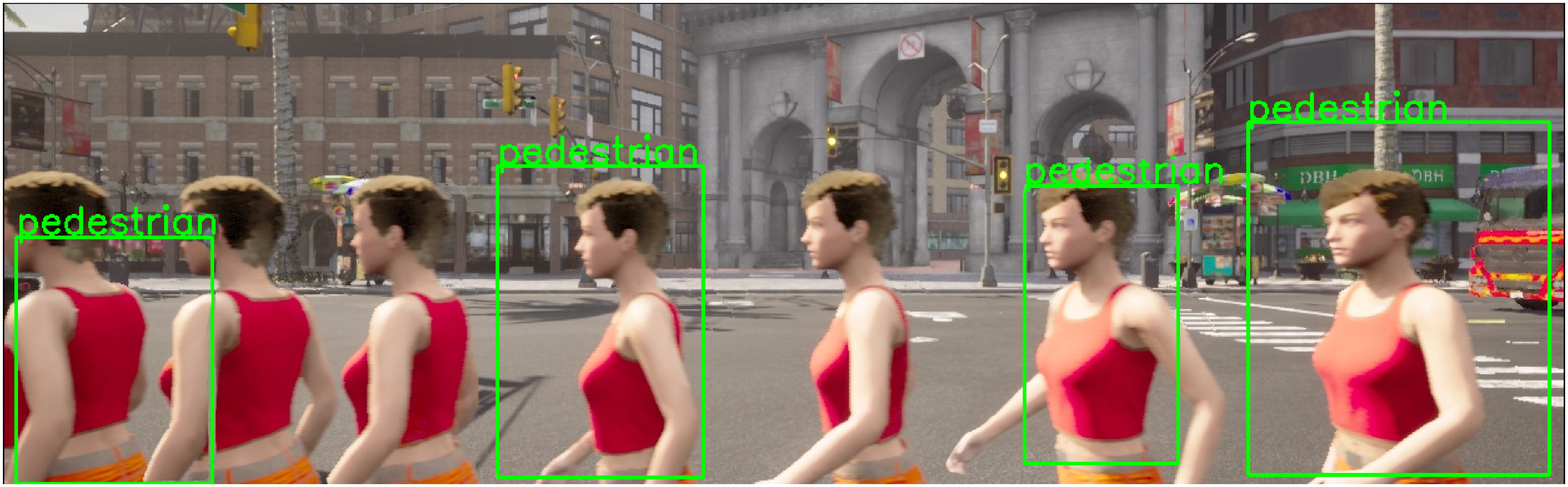}
    \end{subfigure}
    \caption{Comparison between original QD-3DT (top) and our proposed method (bottom). The person in the red shirt is the same person in a different time step.}
    \label{consisitency_fig}
\end{figure}

\begin{figure*}[thpb]
    \centering

    \vspace{10pt}

    \includegraphics[width=0.9\textwidth]{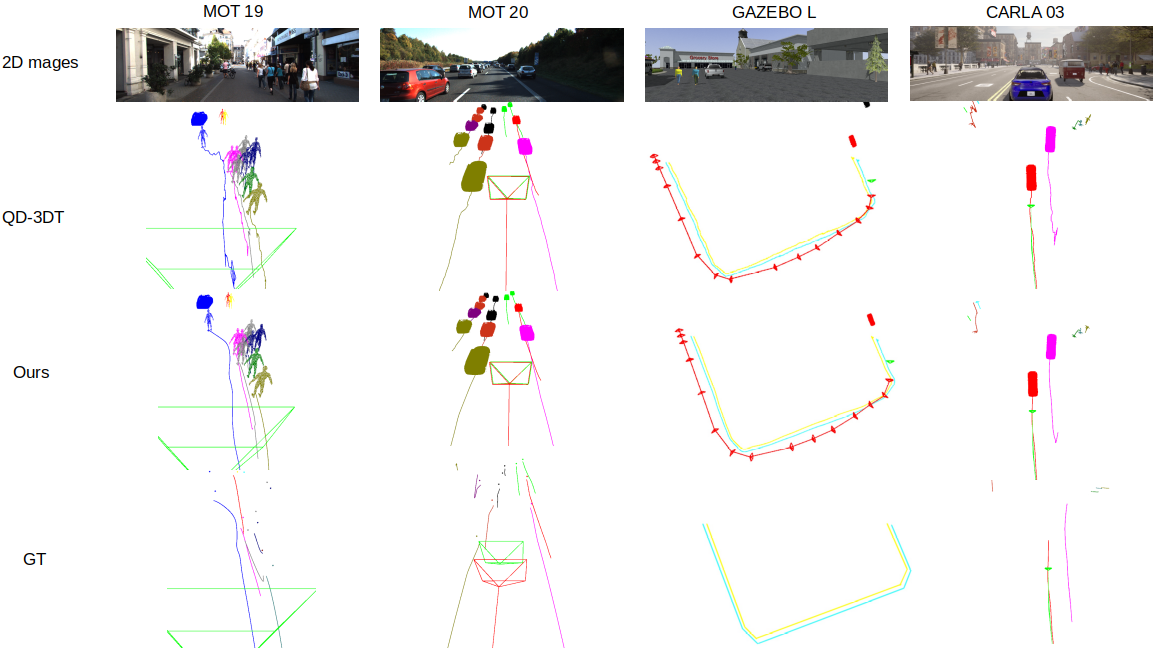}
    \caption{ Trajectories for MOT19, MOT20, GAZEBO L and CARLA03 sequences.  }
    \label{trajeval}
\end{figure*}

As shown in the pedestrian paths of Fig. \ref{trajeval} MOT 19., we compare QD-3DT \cite{hu2022monocular} directly to the proposed system, and the ground truth.
The results demonstrate the high effectiveness of our method by reserving accuracy while removing noise and smoothing trajectories.
We assessed the accuracy of object trajectory estimation by measuring the RMSE of individual trajectories in each frame.
Our results shown in Table \ref{RMSE} demonstrate a significant improvement in trajectory estimation accuracy, achieving at least a 13\% increase in accuracy against the original QD-3DT.

In addition to trajectory analysis, we employe the consistency check.
Table \ref{RMSE} indicates that the QD-3DT based object detection exhibits low precision but high recall values, because many initial detections are incorrect and include duplicates.
Our proposed method increases the accuracy of our detections by getting rid of these duplicates, as visualized in Fig. \ref{consisitency_fig}.
For example, QD-3DT detects the same pedestrian three times, each marked with a 2D bonding box in a different color, while our approach consistently detects the same pedestrian just once.
Moreover, our proposed method does not substantially affect recall values, as it does not introduce new detections or rectify erroneous ones.
Although there is a slight reduction in recall for the CARLA EZL sequence, it is noted that such errors are exceedingly rare, occurring only once across all datasets.

\section{Conclusion}

In this work, we convey a vision based Perception system with a SLAM system which can also track moving objects and optimize and predict the trajectory.

We evaluate the proposed system on both real-world and simulated datasets, which shows great accuracy compared to other methods. Meanwhile, our system provides high performance in optimizing and predicting trajectories and also improves object tracking.

\bibliographystyle{ieeetr}
\bibliography{references}

\end{document}